\documentclass[sigconf]{acmart}

\setcopyright{acmlicensed}
\copyrightyear{2018}
\acmYear{2018}
\acmDOI{XXXXXXX.XXXXXXX}
\acmConference[Conference acronym 'XX]{Make sure to enter the correct
  conference title from your rights confirmation email}{June 03--05,
  2018}{Woodstock, NY}
\acmISBN{978-1-4503-XXXX-X/2018/06}

\usepackage{natbib}
\usepackage{multirow}
\usepackage[table]{xcolor}
\usepackage{arydshln}
\usepackage{xspace}
\usepackage{caption} 
\usepackage{subcaption} 
\usepackage{listings}

\usepackage{pgfcalendar}
\newcount\myjuliandate
\newcount\myjuliantoday
\newcommand{\DaysTo}[3]{%
\pgfcalendardatetojulian{\year-\month-\day}{\myjuliantoday}%
\pgfcalendardatetojulian{#1-#2-#3}{\myjuliandate}%
\advance\myjuliandate by-\myjuliantoday\relax
\the\myjuliandate
}

\definecolor{cylon}{RGB}{224,255,255}  

\newcommand{\methodname}{{GeoGNN}\xspace}

\definecolor{toanblue}{HTML}{8A4B00}

\begin{document}
\title[\methodname: Time Series Geo-Localization]{\methodname: Time Series Geo-Localization using \\ Two-Tower
Graph Neural Networks}

\author{Toan Tran}
\authornote{This work was primarily done during an internship at ORNL.}
\email{vtran29@emory.edu}
\affiliation{
  \institution{Emory University}
  \city{Atlanta}
  \state{GA}
  \country{USA}
}

\author{Waqwoya Abebe}
\email{abebewm@ornl.gov}
\affiliation{
  \institution{Oak Ridge National Laboratory}
  \city{Oak Ridge}
  \state{TN}
  \country{USA}
}

\author{Abhishek Potnis}
\email{potnisav@ornl.gov}
\affiliation{
  \institution{Oak Ridge National Laboratory}
  \city{Oak Ridge}
  \state{TN}
  \country{USA}
}

\author{Supriya Chinthavali}
\email{chinthavalis@ornl.gov}
\affiliation{
  \institution{Oak Ridge National Laboratory}
  \city{Oak Ridge}
  \state{TN}
  \country{USA}
}

\author{Cyrus Shahabi}
\email{shahabi@usc.edu}
\affiliation{
  \institution{University of Southern California}
  \city{Los Angeles}
  \state{CA}
  \country{USA}
}

\author{Li Xiong}
\email{lxiong@emory.edu}
\affiliation{
  \institution{Emory University}
  \city{Atlanta}
  \state{GA}
  \country{USA}
}

\author{Dalton Lunga}
\email{lungadd@ornl.gov}
\affiliation{
  \institution{Oak Ridge National Laboratory}
  \city{Oak Ridge}
  \state{TN}
  \country{USA}
}


\renewcommand{\shortauthors}{Tran et al.}

\begin{abstract}
    This paper investigates a novel concept of time series geolocalization, where the goal is to infer the geographic origin of each raw time series. Successful geolocalization can provide spatial context to time series, enabling downstream location-aware applications. We formalize the problem, adapt core ideas from image geolocalization to establish strong baselines, and propose \methodname, a two-tower architecture. During training, GeoGNN's spatial tower learns embeddings of geographic cell candidates by leveraging the geographic adjacency graph, while the temporal tower extracts informative representations from time series. During inference, each temporal representation is matched against candidate geographic embeddings using dot-product similarity, combined with an auxiliary classification head, to predict the time series' associated geographic origin. Experiments on large-scale, countrywide electricity-consumption datasets demonstrate that \methodname achieves the best performance across datasets and enhances both fine- and coarse-grained geolocalization accuracy by $\sim$27\% on average.
\end{abstract}

\received{20 February 2007}
\received[revised]{12 March 2009}
\received[accepted]{5 June 2009}

\maketitle

\section{Introduction}

Geolocalization refers to the task of determining the geographic location of an entity based on its observed data. This direction has been studied in various domains such as images~\citep{Weyand_2016}, social media posts~\citep{twittergeoloc,rahimi2017}, audio~\citep{kumar2016audio}, and speech~\citep{vanleeuwen2016}. With the advancement of computer vision and rich data availability, image geolocalization has been an active research area with a variety of approaches that can provide reasonable accuracy~\citep{geoclip, jia2025georanker}. Despite extensive study in other modalities, time series -- a cornerstone data type -- remain unexplored. In this paper, we initiate a systematic study of \textit{time series geolocalization}.

Geolocalizing time series is important because geographic context can substantially enrich downstream analysis of time series and improve the reliability of location metadata. Many time series are collected without explicit geographic information due to privacy concerns~\cite{Mao2024DifferentialPF}, technical limitations, or cost constraints~\cite{gpscost}. Even when location metadata is available, it can be unreliable or outdated, as many systems rely on user-reported locations (e.g., Sensor.Community, an open environmental data portal). Associating temporal patterns with their precise geographic origins enables more informed decision-making across domains. For example, in healthcare, geolocalized time series such as admission counts, disease case reports, and prescription fills can enable targeted healthcare resource allocation.  In environmental and mobile application settings, inferring location from usage patterns can enable location-aware monitoring, services, and advertising without requiring explicit GPS permissions.

Furthermore, geolocalization of time series, \textit{if feasible}, will reveal fundamental limitations of a common practice -- location scrubbing as a privacy safeguard. For example, California's  regulatory decisions (CPUC D.11-07-056 and D.11-08-045) mandate strict privacy protections for smart meter data~\citep{westrich2025privacy}. 
However, if time series can be effectively geolocalized from raw temporal signals, this practice may be insufficient, highlighting the need for stronger, privacy-aware data collection and release mechanisms. 


Geolocalizing time series presents several unique challenges. Unlike images, which often contain explicit visual cues such as landmarks, architectural styles, or natural features that are strongly tied to specific places, time series encode location only indirectly through latent, context-dependent temporal patterns. As a result, robust geolocalization of time series requires models that can capture fine-grained, long-range temporal dependencies, and effectively align them with corresponding spatial representations.


To effectively geolocalize time series, we introduce \methodname which is a novel architecture based on two-tower graph neural networks. The temporal tower implements a TimesNet-based backbone, which is a leading architecture for time series classification~\citep{wu2023timesnet}, to extract informative representations from raw time series signals. In parallel, the spatial tower embeds each geographic cell candidate using graph attention networks based on geographical adjacency. Both embeddings are then projected and fused in a two-tower fusion module and combined with an auxiliary classification head. 

It is worth noting that time series geolocalization is fundamentally different from the previous spatio-temporal GNN studies that focus on analysis tasks~\citep{10636792} such as forecasting, imputation, and classification. More specifically, the previous GNNs construct a graph where each node represents a time series entity such as sensors at road segments in traffic forecasting~\citep{Yu_2018}, monitoring stations for air quality prediction \citep{Feng2024SpatioTemporalFN}, and smart meters for load forecasting~\citep{ijcai2021p374}. In contrast, each node in our setting corresponds to a geographical cell that includes many time series entities. Thus, the existing spatio-temporal GNN-based methods are not directly applicable.

To evaluate, we establish a set of baselines by adapting representative
approaches from image geolocalization. Our baselines include a conventional
retrieval-based method, a classification model, an LLM-based retrieval-augmented
generation approach, and a multimodal representation learning method using contrastive
learning. We conduct extensive experiments on large-scale, country-wide electric
power consumption datasets including Spain and the US. Each time series
corresponds to a residential home, commercial building, or industrial facility. \methodname
significantly outperforms the competitive baselines across datasets and evaluation
metrics. On the Spain dataset, \methodname achieves 17.5\% street-level
geolocalization accuracy, compared to the best baseline at 7.4\% and reduces the
median error to 18.3~km in the Northeast US subset, from 51.5~km of the best baseline.
These results demonstrate the effectiveness of our geolocation framework across
diverse geographic settings.

Our contributions can be summarized as follows.
\begin{itemize}
    \item \textit{Novel Problem}. We introduce and define the novel task of time
        series geolocalization, taking the first step toward understanding how
        location can be inferred directly from raw time series signals. To support
        future research, we adapt several strong baselines from the related
        image geolocalization literature.
    \item \textit{Technical Novelty}. We propose \methodname, a two-tower
        architecture that integrates spatial and temporal signals. It employs a graph
        neural network to embed geographic cells and a time series
        backbone to extract temporal features. The two representations are then
        fused to enable accurate geolocation predictions.
    \item \textit{Empirical Evaluation}. Our results demonstrate robust and sustained geolocalization 
        performance on time-series data. \methodname significantly outperforms the
        baselines across multiple datasets and evaluation metrics by a large margin
        up to 64\%. We provide an ablation study and multiple additional analyses to justify 
        our design choices and understand model behavior.        
\end{itemize}

\section{Related Works}

\subsection{Image Geolocalization}
Image geolocalization is an active research area~\citep{Durgam_2024}. \citet{4587784} pioneered image geolocalization at the planet scale by retrieving the most similar images via hand-crafted visual features. This retrieval approach was later expanded to a massive dataset of 35M Flickr images~\citep{carnadallwww}. Meanwhile, \citet{6710175} proposed to use top K nearest neighbors, instead of top 1 match, to improve the system's robustness. With the rapid growth of deep learning, \citet{Weyand_2016} trained convolutional neural networks (CNNs) to classify images into geographical cells. Building on this, the intermediate features from the CNN classifier lead to better retrieval quality than traditional hand-crafted features~\citep{Vo2017RevisitingII}. More recently, transformer-based architectures have replaced early CNNs~\citep{pramanicktransformer, 10203181}. The latest advances leverage powerful pretrained foundation models. \citet{geoclip} applied contrastive learning to align the location embeddings with the pretrained CLIP image embeddings~\citep{radford2021clip}. Following this, \citet{Haas_2024_CVPR} combined multiple technical designs such as semantic geocell creation and multi-task pretraining on top of the pretrained CLIP vision encoder. Furthermore, retrieval-based methods have been enhanced by integrating multimodal LLMs through Retrieval-Augmented Generation (RAG)~\citep{zhongliangimg, jia2024g}. \citet{ghasemi2025geotoken} designed a hierarchical sequence prediction approach that mimics how humans narrow potential locations from coarse to fine levels. Some methods require satellite images~\citep{8578856} or semantic segmentation~\citep{pramanicktransformer} as additional inputs, which are not applicable to time series data. \textit{To construct our baselines, we adapt some representative geolocalization principles while replacing visual encoders by leading time series architectures.}

\subsection{Text- and Audio-based Geolocalization}
For text-based geolocalization, Twitter (now known as X) has been a popular source of data for geolocalization research.  Only a tiny fraction (0.5\% - 1\%) of users opted in to share their location information due to privacy and safety concerns~\citep{Ajao_2015}. This has led to a significant interest of geo-localizing the remaining users based on their tweets as local words and events can be strong indicators of location. An early work \citep{twittergeoloc} employs a simple word frequency-based mechanism to geolocalize users at the city level. Later works with advanced deep learning models, using convolutional neural networks~\citep{Huang_2017}, pretrained embedding models (Doc2Vec and Node2Vec)~\citep{do-twitter}, and transformers~\citep{Lutsai_2024} have further improved the performance for more fine-grained geolocalization. Moreover, audio-based geolocalization is another emerging area of research. \citet{chasmai2025audio} introduced the first global-scale audio geolocalization dataset with a focus on natural sounds. They also demonstrated that methods originally designed for image geolocalization do not yield satisfactory results for audio. This highlights the need to develope modality-specific geolocalization techniques. In our work, we also find that directly applying image geolocalization methods to time series data yields limited performance,  motivating us to develop \methodname.

\subsection{Time Series Analysis}
Time series analysis is a widely studied and popular area in machine learning due to the growing presence of temporal data in modern technologies and real-world systems~\citep{wang2024deeptimeseriesmodels}. Due to its importance, time series analysis has a long history of methodology development ranging from ARIMA~\citep{Anderson1976TimeSeries2E} to recurrent neural networks~\citep{6795963, SALINAS20201181}. These early methods often struggle with information loss when dealing with long sequences. To address this issue, \citet{haoyietal-informer-2021, wu2021autoformer, haoyietal-informerEx-2023, zhou2022fedformer} introduced transformer architectures with self-attention mechanisms, while \citet{liu2022scinet, wang2023micn, wu2023timesnet} leveraged convolutional neural networks with down-sampling and fast Fourier transform. With the rapid advancements in large language models, some studies have explored their use in time series through prompting~\citep{xue2023promptcast, gruver2023large} and fine-tuning~\citep{jin2024timellm, cao2024tempo, changllm4ts}. Recent works aim to provide time series-specialized foundation models that are large-scale pretrained on billions of time series~\citep{ansari2024chronos, goswami2024moment, das2024timefm, ekambaram2024tinytimemixersttms, yao2025towards}. Our work is complementary to these efforts by focusing on the geolocalization task, which requires not only modeling temporal dynamics but also capturing spatial dependencies. We adapt some of the leading time series architectures as baselines and also integrate them into our proposed method as the temporal tower.

\subsection{Spatio-Temporal Graph Neural Networks}
Geospatial information is a valuable modality that has been shown to enhance the performance of time series analysis~\citep{jin2024surveygraphneuralnetworks}. DCRNN~\citep{li2018dcrnn_traffic} was the first to integrate GNNs for time series models by introducing diffusion-based graph convolutions with recurrent units. STGCN~\citep{Yu_2018} captures spatial dependencies by graph convolutions and temporal dynamics by gated temporal convolutions. Subsequently, \citet{Guo_Lin_Feng_Song_Wan_2019} proposed a dual attention mechanism over both time and space that enables the model to learn dynamic importance weights across time steps and nodes. Graph WaveNet~\citep{zonghanwavenet} and MTGNN~\citep{wu2020connecting} learn the graph structure dynamically from the data rather than relying on a fixed, predefined adjacency matrix. \citet{pdformer} employed a graph-informed attention mechanism that integrates spatial structure into the transformer architecture. Inspired by the growing trend of pretraining across various domains~\citep{zhou2023pretrainsurvey}, \citet{li2023generative, li2024opencity} developed pretraining frameworks for spatio-temporal GNNs. Following this, \citet{flashst} fine-tuned pretrained models to tackle distribution shifts, while \citet{jiang2025fstllm} tackled few-shot forecasting by integrating LLMs into spatio-temporal GNNs. Although these STGNNs also combine temporal and spatial information, they address a fundamentally different setting from geolocalization. Particularly, they require the time series to already be assigned to a graph node with known spatial edges, whereas this node/location assignment is exactly what our geolocalization task aims to infer.

\begin{figure*}[!htp]
    \centering
    \includegraphics[width=\linewidth]{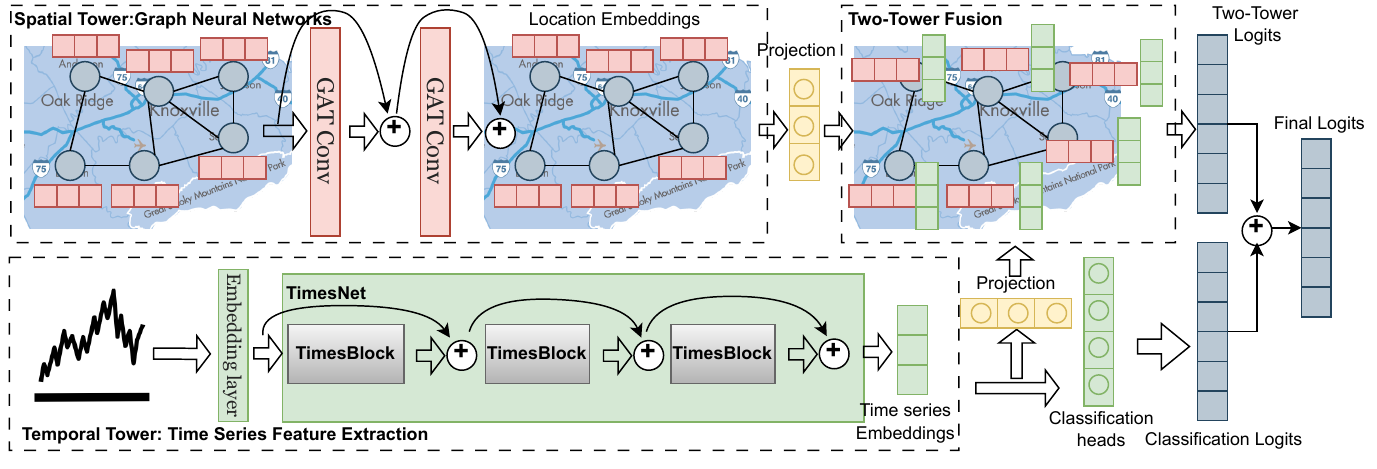}
    \caption{The architecture of \methodname. The spatial and temporal information are separate towers. Subsequently, these two modalities are fused by a two-tower architecture. The output is combined with classification logits to provide the final logits.}
    \label{fig:method}
\end{figure*}

\section{Problem Statement}

We consider the geolocalization problem of inferring the geographic origin of time series. Let $(X_k, y_k)$ denote a sample where $X_k$ is a time series and $y_k = ({lat}_k, {lon}_k)$ represents its ground-truth latitude and longitude of the geographic origin. Given a training dataset $\mathcal{D}_\text{train} = \{ (X_k, y_k) \}$, the goal is to infer the origin location $y$ of an unseen time series $X$.  

We define a mapping function $f_\theta$ which aims to predict the origin location $\hat{y} = f_\theta(X)$ from the input time series. The objective is to learn parameters $\theta$ that minimize the expected geolocalization error over the test distribution.

\begin{equation}
 \displaystyle    \min_\theta \mathbb{E}_{(X, y) \sim \mathcal{D}_\text{test}} \left[ d\left( f_\theta(X), y) \right) \right],
\end{equation}
where $\mathcal{D}_\text{test}$ is an unseen set of time series-location pairs, $d(\hat{y}, y)$ is the Haversine distance between the predicted and ground-truth coordinates.

Due to privacy constraints in most real-world datasets, fine-grained GPS coordinates are often not available. Instead, the location labels are aggregated to high-level regions such as zipcodes or counties~\cite{emami2023buildingsbench, Uria2021smartmeter}. To accommodate this, we consider the feasible space of locations as a discrete, finite set of $N$ non-overlapping geographical cells $\mathcal{C} = {c_1, c_2, \ldots, c_N}$, where each cell $c_i$ is associated with a representative coordinate $({{lat}}_i, {{lon}}_i)$. This relaxes the formulation by assigning each time series label $y_k \in \mathcal{C}$ to one of the predefined cells. The geolocalization task becomes predicting the most likely origin cell $\hat{y} = f_\theta(X) \in \mathcal{C}$.

Despite the granularity constraint, learning to predict geographical cells is a valuable and practical goal. It can provide a foundation for coarse-to-fine hierarchical pipelines by narrowing the search space for future steps of fine-grained localization, clustering, or retrieval. Similar hierarchical approaches have been proven highly effective and exploited in image geolocalization applications~\citep{Haas_2024_CVPR}.

\section{Methodology}
Figure~\ref{fig:method} illustrates \methodname's architecture. 
The spatial tower learns embeddings for each geographical cell by modeling spatial dependencies among locations using a graph neural network. Simultaneously, the temporal tower extracts features from the input time series to produce time-series embeddings. 
The two-tower fusion performs similarity-based matching between the time series and location embeddings. This design allows the model to capture both the intrinsic temporal patterns and the geospatial dependencies of physical potential locations. 

Prior work has explored contrastive learning for image and location alignment~\citep{geoclip}; however, contrastive learning typically requires a large-scale dataset to stabilize its training process with sampling positive and negative pairs. In contrast, our proposed dot-product matching can be more stable and suitable for smaller datasets.

\subsection{Spatial Tower: Graph Neural Network}
The spatial tower aims to learn a rich and geospatially-aware embedding for each geographical cell by explicitly modeling the spatial dependencies between locations. We adopt a graph neural network (GNN) because it naturally represents the geographic domain as a graph, where nodes correspond to cells and edges encode spatial proximity or connectivity. GNNs employ message-passing mechanisms that enable each node to aggregate information from their neighbors. This process allows the model to encode not only the intrinsic location features but also the broader geospatial context. This is critical for our task, as nearby locations are often strongly correlated due to shared environmental conditions, human activity patterns, or infrastructure connectivity. By including such spatial dependencies into the graph structure, GNNs can effectively produce highly informative contextual location embeddings. 

The spatial relationships between $N$ cells are modeled as a graph $G = (V, E)$, where $V = 
\{v_1, v_2, ..., v_N\}$ is the set of predefined, fixed nodes (geographical cells). This set can be formed either directly from datasets with aggregated locations (such as zipcodes and counties) or from spatial partitioning algorithms~\citep{Weyand_2016, Haas_2024_CVPR}. Each node $v_i$ corresponds to a geographical coordinate $({lat}_i, {lon}_i)$.

Let $d(v_i, v_j)$ be the Haversine distance between two nodes $v_i$ and $v_j$. An undirected edge $\{v_i, v_j\}$ is created if the distance between them $d(v_i,v_j)$ is within a predefined radius $\delta$. The set of edges $E$ representing geographical adjacency can be formulated as follows:
\begin{equation}
E = \{\{v_i, v_j\}| v_i, v_j \in V, i \neq j, d(v_i, v_j) < \delta \}
\end{equation}

Once the graph $G$ is constructed, each node $v_i$ is initialized with a learnable feature vector $h_{loc_i}^{(0)} \in \mathbb{R}^{d_0}$. The model employs a stack of Graph Attention (GAT) layers~\citep{velickovicgat} to update the node representations via aggregating information from their connected neighbors. To facilitate training and mitigate over-smoothing, we implement residual connections by adding the input node features to the output of each GAT layer before passing them to the next layer. This process can be represented as the following equation.
\[
h_{loc_i}^{(l+1)} = \text{GATConv}\left(h_{loc_i}^{(l)}, \left\{h_{loc_j}^{(l)}: v_j \in \mathcal{N}_i\right\} \right) + h_{loc_i}^{(l)},
\]
where $l$ is the layer number, $\mathcal{N}_i$ is the set of neighbors of node $v_i$, and the $\text{GATConv}$ function implements the core logic of the Graph Attention Network layer~\citep{velickovicgat}. 
The output of this spatial tower from the final layer (denoted as $\mathcal{L}$) $h_{loc_i}^{(\mathcal{L})}$ is called the location embedding for the location at $v_i$.




\subsection{Temporal Tower: Time Series Feature~Extraction}

Our temporal tower implements the TimesNet backbone~\citep{wu2023timesnet}, which is one of the leading SOTA architectures for general time series analysis. However, this can be seamlessly replaced with any other time series backbones. TimesNet implements a stack of TimesBlocks with residual connections. The TimesBlocks transform time series into a 2D tensor that can represent multi-periodicity patterns by applying Fast Fourier Transform (FFT). The tensor is processed by 2D convolutional filters to effectively capture both the short-term fluctuations within a single cycle (intra-period variation) and the longer-term trends across multiple cycles (inter-period variation). This helps the model develop a robust and comprehensive understanding of the input time series data. Let $X$ denote a raw time series, we project the raw input into a vector by an embedding layer $h_{ts}^{(0)} = \text{Embed}(X)$. The process of the residual stack of the TimesBlocks can be defined as the following equation.

\begin{equation}
h_{ts}^{(l)} = \text{TimesBlock}\left(h_{ts}^{(l-1)}\right) + h_{ts}^{(l-1)},
\end{equation}
where $h_{ts}^{(l)}$ denotes the output of the $l$-th TimesBlock. The detailed operations of TimesNet can be found in the original paper~\citep{wu2023timesnet}. Similar to the spatial tower, the output of the last layer -- $h_{ts}^{\mathcal{L}}$ -- represents the final embedding for time series $X$.


\subsection{Two-Tower Fusion and Logit Computation}
Before fusion, both the location and time series embeddings are projected with linear layers. This helps to enhance the flexibility of our fusion approach. This process can be defined by the following equations.



\begin{equation}
\begin{aligned}
e_{loc,i} &= W_{loc\_proj}\, h_{loc_i}^{(\mathcal{L})} + b_{loc\_proj}, \\
e_{ts}    &= W_{ts\_proj}\, h_{ts}^{(\mathcal{L})} + b_{ts\_proj}.
\end{aligned}
\end{equation}

where $W_{loc\_proj}$, $b_{loc\_proj}$ and $W_{ts\_proj}$, $b_{ts\_proj}$ are trainable parameters that project the embeddings into a shared representation space of equal length.

Our two-tower architecture computes a similarity score between the time series query embedding $e_{ts}$ to each spatial item $e_{loc, i}$. The two-tower logit $z_{tt, i}$  for node $v_i$ is computed as the dot product between the two embeddings:
\begin{equation}
  z_{tt, i} = e_{ts}^\intercal \cdot e_{loc, i}  
\end{equation}

The classification logits $z_{cl}$ are simply calculated by an auxiliary classification head, applied directly to the time series embedding $h_{ts}^{(\mathcal{L})}$. While the fused two-tower logits capture the alignment between time series and location embeddings, this classification head provides an additional direct signal from the time series features alone. This can help the model handle hard time series that have weak spatial dependencies by focusing on their intrinsic temporal patterns.
\begin{equation}
  z_{cl} = W_{cl} h_{ts}^{(\mathcal{L})} + b_{cl},   
\end{equation}
where $W_{cl} \in \mathbb{R}^{N \times d}$ and $b_{cl} \in \mathbb{R}^{N}$ are the trainable parameters of the classification head. Each component $z_{cl,i}$ corresponds to the classification logit for location $v_i$.

The final logit $z_i$ (for location $v_i$) is the sum of these two previous logits:
\begin{equation}
   z_i = z_{tt, i} + z_{cl, i} 
\end{equation}

\subsection{Loss Function and Test Time Inference}
The logit $z_i$ is transformed into a predicted probability $\hat{p}_i$ for location $v_i$ via the softmax function:
\begin{equation}
    \hat{p}_i = \frac{\text{exp}(z_i)}{\sum_j \text{exp}(z_j)}
\end{equation}

Instead of using one-hot labeling, we conduct label smoothing using the Haversine distance from the ground-truth location $y$ to location $v_i$, denoted as $d(y, v_i)$. 

\begin{equation}
p_i = \text{exp}\left( - \dfrac{d(y, v_i)}{\sigma} \right),
\end{equation}
where $\sigma$ is a distance-scale temperature hyperparameter which controls the smoothness of the target distribution.

\methodname implements the Kullback–Leibler (KL) divergence as the loss function. 
\begin{equation}
    \mathcal{L} = D_{KL}(p||\hat{p}) = \sum_i p_i \log \left(\dfrac{p_i}{\hat{p}_i} \right)
\end{equation}

This KL loss, together with the distance-aware label smoothing, encourages the model to make more spatially sensible predictions. Instead of treating all incorrect locations the same, the model learns to favor predictions that are closer to the true location, even if they are not exactly correct. This allows the training process more flexible and robust, especially in regions where nearby locations share similar patterns and are hard to distinguish. 

At the inference stage, the predicted location $f_\theta(X) = \hat{y}$ is the location corresponding to node $v_k$, whose predicted probability is the highest among all nodes. This process can be expressed as the following equation.

\begin{equation}
\begin{aligned}
    f_\theta(X) &= v_k = ({lat}_k, {lon}_k), \\
    \text{where} \quad k &= \arg\max_i \hat{p}_i
\end{aligned}
\end{equation}

\section{Experiments and Results}
\subsection{Experiment settings}
\subsubsection{Datasets.} We conduct experiments using two large-scale, country-wide datasets of electric usage including Spain~\citep{Uria2021smartmeter} and BuildingsBench~\citep{emami2023buildingsbench}. These datasets offer broad geographic coverage and diversity in consumption patterns. The Spain dataset includes nearly 25.5K household-level time series spanning across various regions of Spain, collected from real-world deployed smart meters. Meanwhile, BuildingsBench is a multi-year, multi-institution effort, the largest publicly available power dataset, including 900K synthetic buildings across the US. BuildingsBench was created using building profile simulations and weather data that can capture a wide range of architectural designs, occupancy patterns, and HVAC setups. BuildingsBench was carefully calibrated and validated to ensure strong preservation of true building physics, realistic noise, and operational variability. We consider two scales of BuildingsBench: Northeast US and the entire US. It is worth noting that the locations of these datasets are aggregated at zipcodes for Spain and Public Use Microdata Areas (PUMAs)


\subsubsection{Data Statistics} Each time series entity (i.e., building or household) provides a long-term time series of 1-3 years (Spain) and 1 year (BuildingsBench). We first filter out the entities without geographic origin metadata. While 64\% of the Spain dataset lacks location information, all entities in BuildingsBench include geographic information. We then split into train, validation, and test sets by the time series entities with a ratio of 80:5:15. These long-term time series are then chunked into fixed-length of 2-week windows by default. The data interval is kept the same as the original dataset at 1 hour. All time series windows start with the first event at midnight of Sundays. \textit{By first splitting the train and test sets by entities and then chunking the time series, our pipeline simulates real-world generalization to unseen time series from new entities.}

\begin{figure*}[!htp] 
    \centering 
    \begin{subfigure}[b]{0.29\textwidth} 
        \centering
        \includegraphics[width=\linewidth]{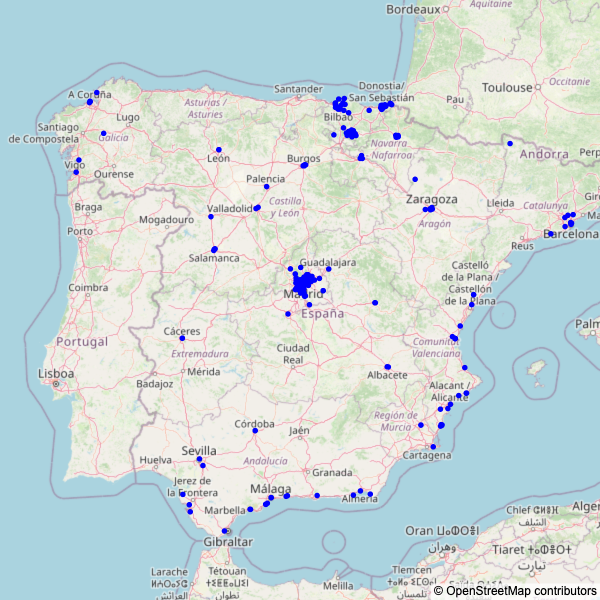} 
        \caption{Spain}
        \label{fig:app-loc-spain}
    \end{subfigure}
    \hspace{1cm} 
    \begin{subfigure}[b]{0.44\textwidth}
        \centering
        \includegraphics[width=\linewidth]{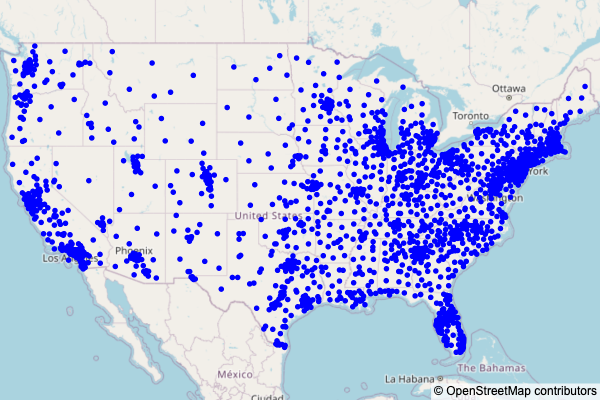} 
        \caption{The US (BuildingsBench).}
        \label{fig:app-loc-us}
    \end{subfigure}
    \caption{Locations of time series in the experimental datasets.}
    \label{fig:app-loc}
\end{figure*}

After processing, there are remaining 6.5K entities, resulting 135K time series windows for training across 305 zipcodes in Spain. Most zipcodes are located in big cities such as Madrid and Bilbao, illustrated in Figure~\ref{fig:app-loc-spain}. Meanwhile, the BuildingsBench dataset offers a much larger scale. In our experiments for BuildingsBench, we use residential data only. More specifically, there are 2M and 11.4M training time series windows, respectively for the Northeast US subset and the full set. In total, there are 2.3K aggregated locations across the US in BuildingsBench, as shown in Figure~\ref{fig:app-loc-us}.

\subsubsection{Baselines.} We adapt several representative and SOTA methods from image geolocalization to establish the baselines.

\textbf{$\bullet$ TimesNet}. Inspired by PlaNet~\citep{Weyand_2016} which formulates geolocalization as a classification problem into discrete geographic cells, we implement the TimesNet~\citep{wu2023timesnet} -- one of the leading SOTA approaches for general time series classification -- as a baseline. The model is trained to predict the location of a time series by classifying it into one of the geographic cells. The coordinate of the predicted cell's centroid is considered as the predicted location.

\textbf{$\bullet$ Retrieval-NN}~\citep{4587784, carnadallwww}. This method retrieves the nearest neighbor time series in the training set and considers the neighbor's location as the predicted location. Instead of using hand-crafted feature extraction as in early works~\citep{4587784, carnadallwww}, we employ a modern retrieval system where a large-scale pretrained time series foundation model performs feature extraction. Particularly, we use the pretrained MOMENT model~\citep{goswami2024moment}.

\textbf{$\bullet$ Retrieval-LLM}~\citep{zhongliangimg}. This approach treats geolocalization as a Retrieval-Augmented Generation (RAG) task by Large Language Models. Motivated by existing RAG approaches with multimodal LLMs for image geolocalization by~\citet{zhongliangimg, jia2024g}, we employ an LLM that receives the input time series and some demonstrations including several nearest and furthest neighbors with their corresponding coordinates to predict the location. More specifically, we employ Llama-3.1-8B-IT~\citep{grattafiori2024llama3herdmodels}. To align with the LLM's tokenizer, which encodes three-digit sequences as single tokens, we represent float numbers using only three decimal digits~\citep{gruver2023large}.

\textbf{$\bullet$ GeoCLIP}~\citep{geoclip}. This approach performs contrastive learning to enforce alignment between image and location embeddings. We replace the pretrained CLIP vision encoder in GeoCLIP with a pretrained time series foundation model. Following the original paper, at the inference stage, the location among all the locations in the training data that has the most similar embedding to the time series is selected as the prediction.

\subsubsection{Evaluation Metrics.}
Following previous studies~\citep{geoclip, Haas_2024_CVPR, zhongliangimg}, we evaluate the geo-localization performance by accuracy at multiple spatial granularities (including 1~km, 10~km, 25~km, 200~km, and 750~km) and Haversine distance error median. These metrics together provide a comprehensive assessment of both fine- and coarse-level localization precision. The accuracy at radius $r$ can be expressed as:


\begin{equation}
    \text{Acc}\  @ \ r\ = \dfrac{1}{|\mathcal{D}_\text{test}|} \sum_i^{|\mathcal{D}_\text{test}|} 1 \left[ d\left(y_i, \hat{y}_i\right) \leq r \right]
    \label{eq:acc}
\end{equation}

\subsubsection{Implementation.} We split the training, validation, and test sets by time series entities, i.e., buildings and houses. By default, each time series is chunked into non-overlapping 2-week windows with hourly intervals, starting at midnight on Sundays. The accuracy metric is then calculated on these windows. This strategy ensures that the models are evaluated on unseen buildings and promotes realistic generalization. All the methods are evaluated on identical train-test split and employ the same early stopping. 

\begin{table*}[!htp]
    \centering
    \begin{tabular}{c|l|ccccc|c}
       \multirow{3}{*}{\textbf{Dataset}}  & \multirow{3}{*}{\textbf{Method}} & \multicolumn{5}{c}{\textbf{Accuracy (\%)} ($\uparrow$)} & \textbf{Distance Error}  \\
        & & (Street) & (Town) & (Metro Area) & (Region) & (Country) & {Median} ($\downarrow$) \\
        &  & \textbf{1 km} & \textbf{10 km} & \textbf{25 km} & \textbf{200 km} & \textbf{750 km} & (\textbf{km}) \\ \hline
        \multirow{6}{*}{Spain} & Random Guess & 1.678 & 6.862 & 9.831 & 36.308 & 97.061 & 312.531 \\
        & TimesNet & 5.793 & 15.130 & \underline{24.842} & \textbf{89.401} & \textbf{99.766} & \textbf{60.332} \\
        & Retrieval-NN & \underline{7.409} & \underline{19.302} & 23.818 & 82.099 & 99.542 & 76.065 \\
        & Retrieval-LLM & 6.747 & 18.560 & 21.795 & 84.104 & \underline{99.687} & 76.044 \\
        & GeoCLIP & 0.148 & 9.190 & 15.438 & 57.527 & 97.711 & 97.366  \\
        & \methodname & \textbf{17.473} & \textbf{28.690} & \textbf{32.820} & \underline{88.578} & {99.645} & \underline{62.667} \\ \hline

        \multirow{5}{*}{BuildingsBench} & Random Guess & 0.249 & 2.113 & 6.094 & 42.764 & 94.745 & 250.825 \\
        \multirow{5}{*}{(Northeast US)} & TimesNet & \underline{9.561} & \underline{22.008} & \underline{36.962} & \underline{73.461} & 97.408 & \underline{51.497} \\
        & Retrieval-NN & 1.573 & 6.329 & 13.720 & 53.498 & 96.222 & 175.629 \\
        & Retrieval-LLM & 0.916 & 3.932 & 11.279 & 54.630 & \underline{99.045} & 173.164 \\
        & GeoCLIP & 1.000 & 7.511 & 12.309 & 47.795 & 95.589 & 216.691\\
        & \methodname & \textbf{13.372} & \textbf{36.918} & \textbf{57.753} & \textbf{90.485} & \textbf{99.305} & \textbf{18.339} \\ \hline 

        \multirow{5}{*}{BuildingsBench} & Random Guess & 0.042 & 0.102 & 0.356 & 3.564 & 20.415 & 1511.607 \\
        \multirow{5}{*}{(US)} & TimesNet & \textbf{14.155} & \textbf{17.926} & \underline{30.690} & \underline{67.164} & \underline{84.301} & \underline{66.486} \\
        & Retrieval-NN & 1.000 & 1.632 & 3.903 & 18.201 & 48.336 & 784.933 \\
        & Retrieval-LLM & 0.608 & 1.230 & 3.276 & 17.036 & 49.157 & 767.195 \\ 
        & GeoCLIP & 0.418 & 0.643 & 1.276 & 7.914 & 38.695 & 971.668 \\
        & \methodname & \underline{11.170} & \underline{17.731} & \textbf{32.470} & \textbf{74.861} & \textbf{90.188} & \textbf{59.411} \\ \hline
    \end{tabular}
    \caption{Performance comparison of \methodname against baselines across multiple datasets and metrics. Bold indicates the best result, while underlined values denote the second best.}
    \label{tab:overall}
\end{table*}

\subsection{Overall Evaluation}
Table~\ref{tab:overall} presents the overall evaluation results of the methods on three datasets. The conventional classification and retrieval approaches offer reliable performance and significantly outperform the random guess baseline. This indicates the strong correlation between the time series and their corresponding locations. Retrieval-NN performs relatively well, achieving the second-best street-level localization accuracy of 7.4\% on the Spain dataset. However, their performance degrades dramatically when applied to the larger datasets. As the dataset grows, the feature space becomes increasingly crowded with similar patterns from different locations. This can hurt the quality of top-1 similarity-based retrieval. While \citet{zhongliangimg} demonstrate multimodal LLMs, especially GPT4-V~\citep{gpt4v}, notably boost image geolocalization precision of the retrieval-based approach. GPT4-V's broad training on web-scale image-caption pairs helps gain a general understanding of visual scenes and global contexts, but the current capabilities of LLMs for time series are still limited~\citep{tan2024are}. As a result, integrating LLMs for time series (referred to as Retrieval-LLM) has little to zero benefit over the standard nearest neighbor retrieval method.

Although GeoCLIP demonstrated its strong performance for image geolocalization~\citep{geoclip}, for time series, GeoCLIP achieves better utility compared to random guess but notably underperforms compared to the conventional approaches such as the TimesNet classifier and retrieval-based methods. Given the difference between image geo-localization's and our problem's settings that the locations of time series are aggregated at geographical cells, GeoCLIP is ineffective at aligning these high-frequency latitude and longitude coordinates' and their corresponding time series' embeddings. \methodname offers robust geo-localization precision across the datasets and granularities with the highest accuracy in most cases. Especially, \methodname enhances the accuracy at street level by 2.5 times for the Spain dataset and reduces the distance error median by 64\% for the NorthEast dataset compared to the best baseline's performance.

\subsection{Ablation Study}
Table~\ref{tab:ablation} presents the results of our ablation study on the Northeast US dataset. The first row indicates the full \methodname, where all the proposed technical designs are included. In this study, we remove key components of \methodname to assess their individual contributions to the  final performance. 

First, we examine the impact of the proposed Kullback-Leibler (KL) divergence loss by replacing it with the standard cross-entropy (CE) loss, as used in PlaNet~\citep{Weyand_2016}. The CE loss punishes the model based on whether the prediction matches the single positive label (i.e., the ground-truth location), without considering other locations. This helps to improve the model's ability to distinguish nearby locations, increasing the street-level accuracy by 3\%. Meanwhile, the KL loss treats all locations as potential labels and weights them based on their physical distances to the ground-truth location. This encourages the models to understand the spatial relationships among all locations, leading to better performance at coarser granularities. 


Subsequently, we evaluate the role of the two-tower architecture by removing the spatial tower and fusion module. The remaining architecture includes only the temporal tower using TimesNet, followed by the classification head. This model is trained with the KL loss. Removing the two-tower GNN component leads to the largest performance decline across all levels. Notably, with the two-tower architecture, \methodname reduces the distance error median by nearly 33\% compared to the single-tower version. Compared to the TimesNet in the baseline which was trained by CE loss, training with KL loss significantly reduces the median error by around~41\%.

Finally, we ablate the classification head. While removing the classification head does not cause a significant performance drop, we observe a modest decline in accuracy at all granularity levels and slight increase of the distance error median by about 2\%. Overall, these results validate the importance of the proposed modules in \methodname. The KL loss guides the model to consider physical distances between negative-label locations, while the GNN two-tower architecture sustainably contributes to the final performance.

 
\begin{table*}[!htp]
    \centering
    \begin{tabular}{l|ccccc|c}
    \multirow{3}{*}{\textbf{Ablation}} & \multicolumn{5}{c}{\textbf{Accuracy (\%)} ($\uparrow$)} & \textbf{Distance Error}  \\
         & (Street) & (Town) & (Metro Area) & (Region) & (Country) & {Median} ($\downarrow$) \\
         & \textbf{1 km} & \textbf{10 km} & \textbf{25 km} & \textbf{200 km} & \textbf{750 km} & (\textbf{km}) \\ \hline
         \methodname & {13.372} & \textbf{36.918} & \textbf{57.753} & \textbf{90.485} & \textbf{99.305} & \textbf{18.339} \\         \hdashline
        w/o KL Loss & \textbf{16.649} & 33.404 & 54.763 & 89.499 & 98.820 & 20.721 \\
        w/o Two-Tower GNN & {12.620} & 28.676 & 47.174 & 83.161 & 98.446 & 27.953 \\
        w/o Classification Head & 12.644 & 36.226 & 57.202 & 90.263 & 99.240 & 18.746  \\
    \end{tabular}
    \caption{Ablation study on the impact of \methodname's components. Bold values indicate the best performance.}
    \label{tab:ablation}
    \vspace{-0.5cm}
\end{table*}

\subsection{Findings \& Analyses}
For consistency, all experiments in the Findings 
\& Analyses section are conducted on the Northeast US dataset.
\subsubsection{GNN versus GeoCLIP's lat-lon encoder.}
To demonstrate the effectiveness of GNN for location embeddings, we replace the GNN by an existing lat-lon encoder, used in GeoCLIP~\citep{geoclip}. We also remove the classification head, so the accuracy is fully dependent on the two-tower logits. GeoCLIP's lat-lon encoder transforms the latitude and longitude coordinates into Fourier features. These features are then passed through a Multi-Layer Perceptron (MLP) to produce a location embedding. Figure~\ref{fig:location-encoder-acc}~(left) depicts the accuracy of \methodname using the GNN and lat-lon encoder. The GNN-based one achieves higher accuracy at all granularities with a large gap of 21\% at 25~km. By aggregating information across neighboring locations, the GNN allows the embeddings to reflect from itself and its neighbors, and enables the construction of multi-scale relationships through message passing. Meanwhile, the GeoCLIP's lat-lon encoder encodes each single point separately by leveraging the raw coordinate with little awareness of the surrounding structure.

\subsubsection{Effects of Distance-Scale Temperature.}
\begin{figure}[!htp]
    \centering
    \includegraphics[width=0.48\linewidth]{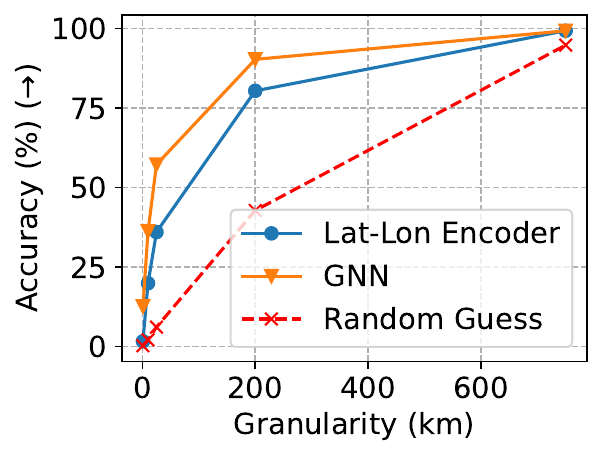}
    \includegraphics[width=0.48\linewidth]{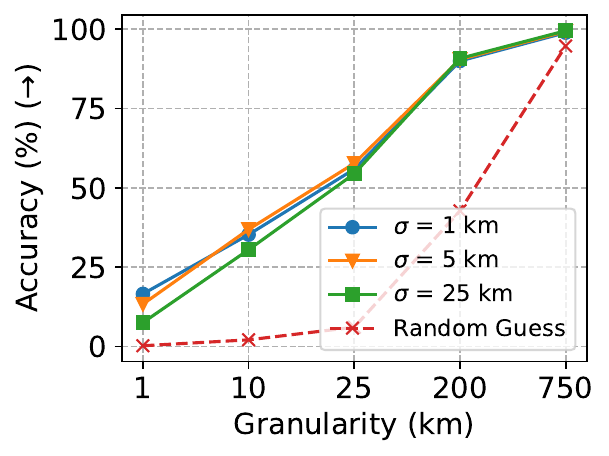}
    \caption{Studies on the location encoders (left) and distance-scale temperature (right).}
    \label{fig:location-encoder-acc}
\end{figure}

The distance-scale temperature is a crucial hyperparameter that modulates the sensitivity of spatial distances. Figure~\ref{fig:location-encoder-acc}~(right) presents the results of various values of $\sigma$ at 1, 5, and 25~km. A smaller value of $\sigma$ (at 1~km) is generally better for street-level localization as it can penalize the smaller distance error. Meanwhile, the value of 5~km offers better performance on town- and metro-area-level accuracy. The large configuration of 25~km can face over-smoothing issues that decrease the fine-level performance but enhance the region- and country-level accuracy.

\subsubsection{Time Series Length.}
We retrain \methodname using time series with various lengths -- 1~week, 2~weeks, and 3~weeks while keeping the same hourly intervals. As shown in Figure~\ref{fig:length-topk}~(left), the results follow a consistent trend. The longer time series, the higher the accuracy of geolocalization, especially at the fine granularities from 1~km to 25~km. Intuitively,  longer time series carry more informative patterns that enhance the overall performance. Notably, the street-level accuracy is improved from 11.8\% (1 week) to 17.3\% (3 weeks) -- a gain of nearly 50\%. This suggests that short-term sequences tend to exhibit similar patterns across different locations, which makes it harder for the model to identify the location. At coarser scales (200~km and 750~km), the returns are diminishing, as regional-level localization requires only broad, general patterns, which can already be captured in shorter sequences.

\subsubsection{Top-K predictions.}
\begin{figure}[!htp]
    \centering
    \includegraphics[width=0.485\linewidth]{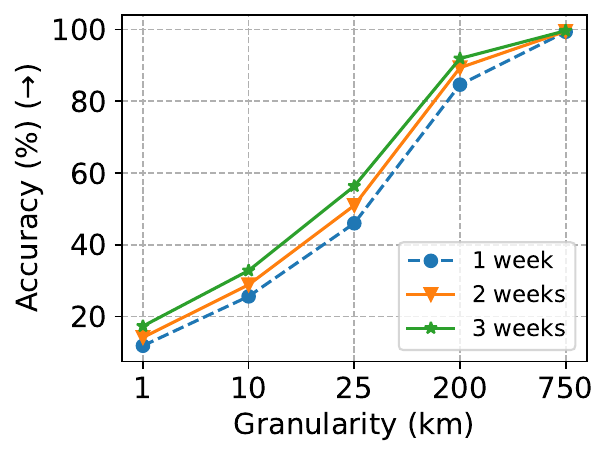}
    \includegraphics[width=0.505\linewidth]{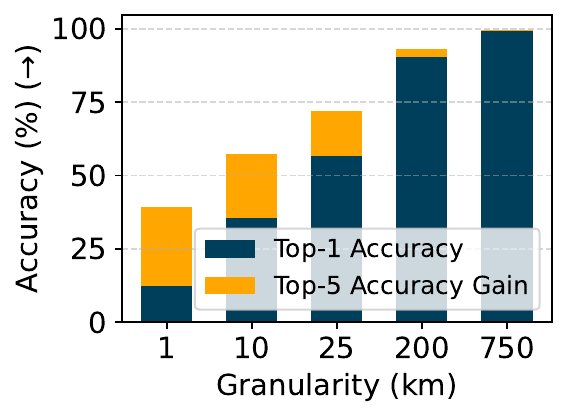}
    \caption{Analyses on the time-series length (left) and top-K predictions (right).}
    \label{fig:length-topk}
\end{figure}

We examine the top-K predictions of \methodname. Figure~\ref{fig:length-topk}~(right) depicts the top-1 and top-5 accuracy. The gains are significantly large at fine granularities but vanish at coarse levels. Notably, at the street-level, top 5 predictions boost the accuracy by three times to around 39\%. Therefore, even when the most confident prediction is incorrect, the true prediction is usually among the highest-confident choices. This suggests the strong localization capability of \methodname with some uncertainty in probability ranking. 

\section{Conclusion}
This work takes a first step toward understanding whether and how the geographic origin of a time series can be inferred from the signal alone. 
We have introduced, formalized, and designed \methodname to effectively geolocalize time series.
Through extensive experiments, \methodname demonstrates robust performance and significantly outperforms the baselines across datasets and evaluation metrics. Our results demonstrate that time-series geolocalization is substantially feasible. This direction opens the door to integrating geolocalization to enable downstream location-aware applications and calls into question the common privacy practice of relying on location scrubbing as a safeguard, motivating research into stronger privacy-preserving data release techniques.

However, one limitation of our work is that the locations are aggregated rather than exact locations, due to limited data availability. Nonetheless, predicting aggregated cells serves as a valuable foundation for finer-grained geolocalization and can naturally integrate with existing hierarchical frameworks developed for image-based localization~\citep{Haas_2024_CVPR}. Additionally, our experiments primarily focus on power consumption data; other types of time series may carry different levels of spatial information. We leave this for future work. We believe this work opens up exciting new directions for time series representation learning and location-aware applications, while raising important questions around location privacy in real-world sensing systems.


\bibliographystyle{ACM-Reference-Format}
\bibliography{ref} 

\appendix
\appendix

\section{Baseline Implementation}

\subsection{Random Guess}
\begin{figure}[!htp]
    \centering
    \includegraphics[width=0.5\linewidth]{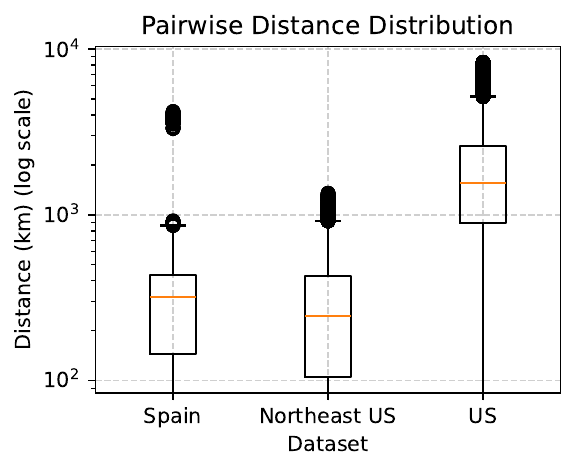}
    \caption{Pairwise Distances among feasible locations across the datasets}
    \label{fig:app-dist-box}
\end{figure}

\begin{figure*}[!htp]
    \centering
    \begin{subfigure}[b]{0.32\textwidth}
        \centering
        \includegraphics[width=\linewidth]{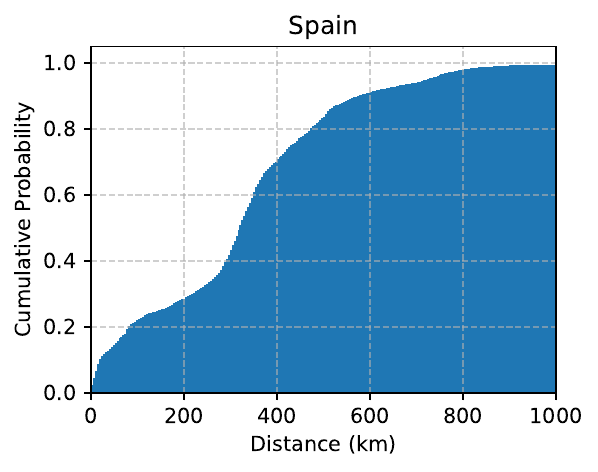}
        \caption{Spain}
    \end{subfigure}
    \hfill
    \begin{subfigure}[b]{0.32\textwidth}
        \centering
        \includegraphics[width=\linewidth]{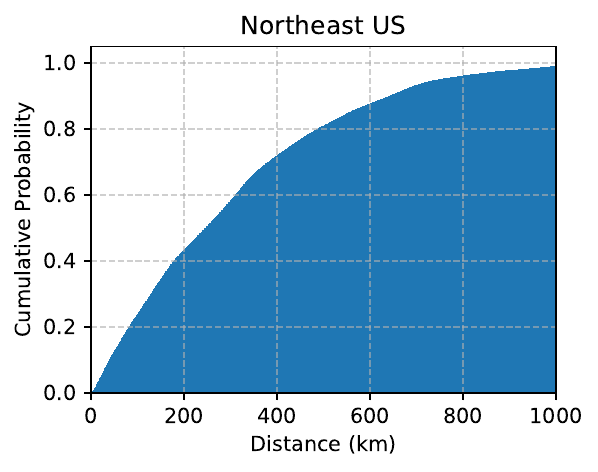}
        \caption{Northeast US}
    \end{subfigure}
    \hfill
    \begin{subfigure}[b]{0.32\textwidth}
        \centering
        \includegraphics[width=\linewidth]{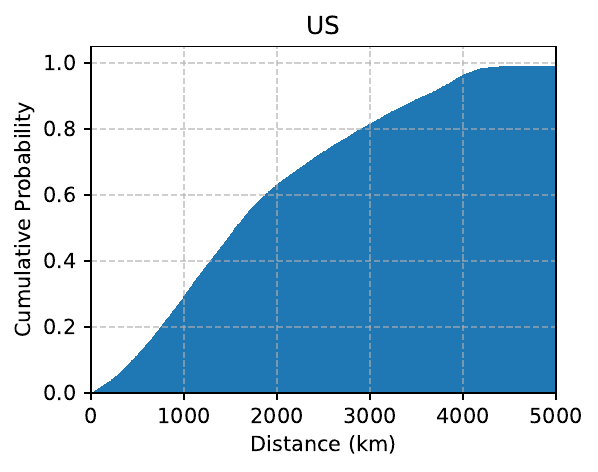}
        \caption{US}
    \end{subfigure}
    \caption{Cumulative probability distribution of pairwise distances among feasible locations.}
    \label{fig:app-dist-prob}
\end{figure*}

We implement a random guess baseline to form the lower bound for the task. Given the set of all feasible locations, the model uniformly samples a location as the prediction. Figure~\ref{fig:app-dist-box} provides a comparison of the pairwise distances among all the feasible locations. The wide geographic spread indicates the difficulty of our geolocalization task. Particularly, the NorthEast US dataset has the smallest pairwise distances on average and fairly close to the Spain's distribution. Meanwhile, the US-wide BuildingsBench exhibits significantly larger pairwise distances. This reflects a more complex and challenging task due to the broader geographic scope and greater spatial dispersion. We provide the detailed cumulative probability distributions of pairwise distances in Figure~\ref{fig:app-dist-prob}.

\subsection{TimesNet}
We implement TimesNet by reusing the code from the original implementation\footnote{https://github.com/thuml/Time-Series-Library}. The hyperparameters are configured by the default recommended by the library, provided in Table~\ref{tab:app-timesnet}. The number of classes are the number of unique locations that come from the dataset. For fair comparison, we also employ the same TimesNet configuration for the temporal tower of \methodname. The optimizer and early stopping strategy are also identical to \methodname's implementation, which is described in the section below. The final performance is evaluated by the checkpoint that performs the best on the validation set.

\begin{table}[!htp]
    \centering
    \begin{tabular}{lc}
       \textbf{Parameter}  & \textbf{Value}  \\ \hline
       Number of encoder layers  & 2  \\
       Dimension of feed-forward network & 32 \\
       Number of dominant frequency components & 3 \\
       Number of 2D kernels & 6 \\
       Number of input features & 1 \\
       Dropout & 0.1 \\
       Dimension of input embeddings & 16 \\
       Number of output channels & 7       
    \end{tabular}
    \caption{TimesNet's hyperparameters}
    \label{tab:app-timesnet}
\end{table}

\subsection{Retrieval-NN}
This method predicts the location of a time series query by retrieving the most similar sample from the training set. The predicted location is the one corresponding to the retrieved sample. We implement a modern retrieval approach. We employ a pretrained time series foundation model that can produce generalizable representation due to its large-scale and broad pretraining. Particularly, we implement the MOMENT model (small)~\citep{goswami2024moment}. MOMENT is an encoder-decoder architecture, pretrained with masked time series reconstruction on 1.23 billion timestamps from multiple domains.

For retrieval, we use only the encoder from the pretrained encoder-decoder model. Each time series is passed through the encoder to obtain a fixed-length embedding. For efficiency, we employ \texttt{FAISS}\footnote{https://github.com/facebookresearch/faiss} to index all the training embeddings. At inference time, the query embedding is compared against this index using the standard Euclidean distance:

\[ {L}_2(z_q, z_i) = || z_q - z_i ||_2 = \sqrt{\sum_{j=1}^d \left(z_{q,j} - z_{i,j}\right)^2}, \]

where $z_q$ is the query embeddings, $z_i$ is the embeddings of the $i$-th training time series, and $d$ is the embedding dimension. At the inference time, the prediction is the corresponding location of the sample that has the minimum l2 distance to the query:

\[ \text{pred} =  \text{loc}_k, \; \text{where} \; k = \arg\min_i ||z_q - z_i||_2 \]

This retrieval system implementation is fast and scalable to efficiently search nearest neighbors over millions of time series. It also benefits directly from the generalization ability of the foundation model without requiring task-specific fine-tuning.

\subsection{Retrieval-LLM}
This approach is built on top of the standard retrieval method (i.e., Retrieval-NN). Instead of considering the top 1 retrieval output directly as the prediction, Retrieval-LLM retrieves top K samples from the training set and uses them as in-context demonstrations for LLMs. More specifically, we retrieve top 3 both nearest and furthest neighbors as demonstrations, as inspired by previous works for image geolocalization~\citep{zhongliangimg}. While the nearest neighbors help the LLM understand what plausible matches look like, the furthest neighbors by contrast guide the LLM away from incorrect areas. Our implementation uses the popular Llama-3.1-8B-IT model~\citep{grattafiori2024llama3herdmodels}, as it offers a good balance between the computational cost and generation quality. The prompt example is provided as follows.
\begin{lstlisting}[language=,basicstyle=\ttfamily\small,numbers=none,columns=fullflexible,keepspaces=true,breaklines=true,breakatwhitespace=true]
<|begin_of_text|><|start_header_id|>system<|end_header_id|>
You are a helpful assistant that geolocates a given hourly power consumption time series of two weeks. Each data point is a power consumption value in kW for each hour, starting from 00:00 on Sunday. You are given the time series, and the locations of the most similar and most dissimilar time series. Your task is to reason step-by-step and predict the latitude and longitude of the given time series.

At the end of your reasoning, output ONLY the result in the following JSON format, DO NOT include any other text after the JSON. The result must be a valid JSON object following the schema:
```json
{
  "latitude": float,
  "longitude": float
}

<|start_header_id|>user<|end_header_id|>
The time series is: 
[0.256, 1.023, 0.985, 0.654, ...]
For your reference:
The most similar time series are at: (41.584, -74.173), (38.719, -77.634), (42.186, -71.346)
The most dissimilar time series are at (41.582, -87.948), (38.310, -122.560), {29.769, -95.101}

<|start_header_id|>assistant<|end_header_id|>
\end{lstlisting}

\vspace{\baselineskip}

For efficiency, we implement the index search by \texttt{FAISS} and the LLM inference by \texttt{vLLM}\footnote{https://github.com/vllm-project/vllm}. The LLM generation configuration is set as default by \texttt{vLLM}, employing the top-p sampling strategy with $p = 0.95$, the temperature of 1.0, and no penalty for repetition.

\subsection{GeoCLIP}
GeoCLIP~\citep{geoclip} is a contrastive learning framework, designed for image geolocalization. The original model employs a pretrained CLIP-based vision encoder for image embeddings and introduces a novel location encoder for coordinate embeddings. In our adaptation, we replace the original vision encoder with a pretrained encoder from a foundation model specialized for time series. More specifically, we use MOMENT -- the same as the one implemented for our retrieval-based approaches. We reuse the original GeoCLIP's codebase\footnote{https://github.com/VicenteVivan/geo-clip/tree/main} to implement the location encoder. The location encoder remains unchanged and continues to learn a trainable mapping from coordinates into the same embedding space.

During training, the model optimizes the InfoNCE loss:
\[ \mathcal{L}_{\text{InfoNCE}} = -\dfrac{1}{N} \sum_i \ln \dfrac{e^{v_i \cdot w_i}/ \tau}{\sum_j e^{v_i \cdot w_j / \tau}} - \dfrac{1}{N} \sum_j \ln \dfrac{e^{v_j \cdot w_j / \tau}}{\sum_i e^{v_i \cdot w_j / \tau}}, \]

which encourages the dot product similarity between matching time series and location pairs $(v_i \cdot w_i)$ to be high, while penalizing the similarity between non-matching pairs. The temperature $\tau$ is set at 0.07 as recommended by the original CLIP paper~\citep{radford2021clip}. The optimizer and early stopping strategy are reused from the implementation of TimesNet \& \methodname.

At inference time, we follow the original procedure in the original GeoCLIP. A query time series is embedded by the time series encoder. The predicted location is then determined by identifying the location in the training set whose embedding has the highest similarity to the query embedding. Particularly, we apply the cosine similarity as the matching criterion:

\[ \text{pred} = \text{loc}_k, \; \text{where} \; k = \arg\max_i \left( \dfrac{z_q \cdot w_i }{ ||z_q||_2 \cdot || w_i||_2} \right) \]

\section{\methodname}
We implement \methodname using PyTorch. For the spatial tower (GNN), the graph construction is implemented by \texttt{sklearn.neighbors}. Given the difference of geographic distribution on Spain and the US. We configure the graph-construction radius $\delta$ at 3.0~km and 10.0~km, respectively for Spain and the US. The embedding layer transforms each location index to a 512-dimensional dense vector. These vectors are then processed by the \texttt{GATConv} layers with residual connections. We implement two GAT layers with \texttt{torch\_geometric}. The first layer projects the node features into a higher-dimensional space using 8 attention heads. The outputs are concatenated and fed through an ELU activation and dropout for regularization. The second GAT layer processes the input using only a single attention head. For the temporal tower (TimesNet), we reuse the implementation from the previous TimesNet baseline's implementation. Both the spatial and temporal embeddings are then projected into a shared 512-dimension representation space before fusing by the dot product. For the KL loss function, we apply a distance-scale temperature $\sigma$ of 5~km.

For training, we employ the standard AdamW optimizer with a weight decay of 0.01 for L2 regularization that helps to prevent overfitting by penalizing large weights. We use the learning rate of 0.001 for Spain and 0.003 for the US. All deep learning-based methods including TimesNet, GeoCLIP, and \methodname implement the same optimizer setting. We also apply an early stopping strategy with a patience of 5 validation steps. The model is evaluated on the checkpoint that performs the best on the validation set. For all the datasets and methods, i.e., TimesNet, GeoCLIP, and \methodname, we set a batch size of 1024 and train for 20 epochs.




\end{document}